\documentclass[journal]{IEEEtran}
\usepackage{graphicx}
\usepackage{amsmath}
\usepackage{amssymb}
\usepackage[algo2e]{algorithm2e}
\usepackage[tight,footnotesize]{subfigure}
\usepackage{multicol}

\usepackage{url}
\usepackage{array}
\usepackage{epsfig}
\usepackage{algorithm}
\usepackage{subfigure}
\usepackage{multicol}
\usepackage{color}
\usepackage{epstopdf}
\usepackage{float}

\def \xt {{\bf x}^{(t)}}
\def \x  { { \bf x}}
\def \Dt { { \bf D^{(t)}}}

\def \D { { \bf D}}

\def \v { { \bf v}}
\def \u { { \bf u}}
\def \p { { \bf p}}
\def \pt {{\bf p}^{(t)}}
\def \W { { \bf {\mathcal W}}}
\def  \A { \mathcal F_u}
\def \y {{\bf y}^{(t)}}
\def \y { { \bf y}}
\def \yt { { \bf y}^{(t)} }
\def \s   { { \bf {s}}}
\def \z   { { \bf {z}}}
\def \R{\mathcal R}
\def \P {\mathcal P}

\def \N {\mathcal N}

\def \X {{\bf X}}
\def \alphaj {\boldsymbol{\alpha}_j}

\def \kkk{\color{black}}
\ifCLASSINFOpdf

\else

\fi

\begin{document}

\title{Dependent Nonparametric Bayesian Group Dictionary Learning for online reconstruction of Dynamic MR
images }

\author{Dornoosh ~Zonoobi,~\IEEEmembership{Member,~IEEE,}
        Shahrooz~Faghih~ Roohi,~\IEEEmembership{Member,~IEEE,}
        and~Ashraf~A.~Kassim,~\IEEEmembership{Senior~Member,~IEEE}
\thanks{Authors are  with the Department
of Electrical and Computer Engineering, National University of Singapore, Singapore}
\thanks{Asterisk indicates corresponding author. e-mail: dornoosh@nus.edu.sg}}

%
\markboth{IEEE Transactions on Biomedical Engineering,  Jan~2015}%
\maketitle

\maketitle

\begin{abstract}

In this paper, we introduce a dictionary learning based approach applied to the problem of real-time reconstruction of MR image sequences that are highly undersampled in k-space. Unlike traditional dictionary learning, our method integrates both global and patch-wise (local) sparsity information and incorporates some priori information into the reconstruction process. Moreover, we use a  Dependent Hierarchical Beta-process as the prior for the group-based dictionary learning, which adaptively infers the dictionary size and the sparsity of each patch; and also  ensures that similar patches are manifested in terms of similar dictionary atoms. An efficient numerical algorithm based on the alternating direction method of multipliers (ADMM) is also presented. Through extensive experimental results we show that our proposed method achieves superior reconstruction quality, compared to the other state-of-the- art DL-based methods.
\end{abstract}


\IEEEpeerreviewmaketitle

\section{Introduction}


Magnetic Resonance Imaging (MRI) is essential  noninvasive tool for  visualization and diagnosis of   the anatomy and function of the body.
It is however burdened by its intrinsic slow data acquisition process. Since data acquisition is sequential in MR imaging
modalities, the scan time (time to get enough data to accurately
reconstruct one frame) is reduced if fewer measurements
are needed for reconstruction. The main goal of much research efforts is
to be able to use a smaller set of samples than normally
required to reconstruct the original images. However, when $k$-space
is under sampled, the Nyquist criterion is violated, and
conventional Fourier reconstructions exhibit aliasing artifacts.


 Compressed sensing (CS) \cite{Candes20062,Candes2006,Candes2008,Donoho}, on the other hand,  has been shown to be able to overcome these challenges and recover MRI images from much smaller k-space measurements than conventional reconstruction methods. To achieve this, earlier CS-based methods assumed that the MRI images have a sparse representation in some known transform domain such as Wavelets  \cite{Lustig2007,Haldar2011,Majumdar2011,NiuYan2008} Contourlet \cite{Gho2010}, finite difference domains \cite{Trzasko2011}.



Most of the prior work in compressed sensing MRI has been based on pre-constructed sparsifying transform.  Even though pre-defined transformations are easier to implement,  it was shown that such transforms sometimes lead to an insufficient and over-simplistic  sparse representation of images which can usually capture only one feature of
the image.   For example, the widely used 2D Wavelets is able to recover point-like features but fails to capture geometric regularity such as smooth contours,  Contourlets can sparsely represent curve-like features  but not the points in images  and finite difference may lead to staircase artifact \cite{Qu2013}.  Therefore, compressed sensing MRI (CSMRI) with nonadaptive, global sparsifying transforms, is usually limited in typical MR images to 2.5-3 fold undersampling \cite{Ravishankar2011}.

To tackle the problems of these global transforms, some recent methods enforced the sparsity  on image subregions called {\it patches}. The shift  from using global image sparsity to patch-based sparsity
is crucial in capturing local image features and has been shown to potentially remove noise and aliasing artifacts \cite{Qu2012,Qu2013}.
An example of such methods is the patch-based directional wavelets method (PBDW) \cite{Qu2012} in which authors  proposed to rearrange the pixels of each patch parallel to the geometric direction of that patch and then perform the Wavelet transform. More recently, \cite{Qu2013} proposed to use  {\it block matching} to group similar patches together and then performed  3D Haar wavelet transform on each stack of similar patches. The proposed approach is shown to achieve greater sparsity than the conventional global Wavelet transform due to the similarity of the patches in each group. Although these methods avoided some of the problems associated with pre-defined representations,  they still lack the ability to contain a variety of underlying features, such as edges and textures.

As an alternative, finding (learning) an adaptive patch-level basis (called {\it dictionary}), specifically tailored to  the image under consideration, is shown to yield significantly enhanced  performance in MR image reconstruction \cite{Aharon2006,Ravishankar2011,Huang2014}.
Dictionary Learning (DL) has been applied to MR images as a sparsifying basis for reconstruction  (e.g., K-SVD LOST\cite{Akccakaya2011} and DLMRI \cite{Ravishankar2011}). The major problem with these methods also known as {\it parametric} DL  methods, is their dependence on the initial value of patches' sparsity level and number of dictionary atoms. When these settings do not agree with ground truth, the performance can significantly degrade.
To avoid this, \cite{Lin2013} applied a Beta-bernoulli process as a nonparametric Bayesian method for adaptive learning of dictionaries. Their work,  based on {\it Beta Process Factor Analysis} (BPFA) \cite{Paisley2009}, is utilized in  \cite{Zhou2012}  for image in-painting.  \cite{Huang2014} also applied  BPFA in combination with a total variation regularization term to reconstruct MR images from undersampled $k-$space. Another method,  {\it dependent Hierarchical Beta Process }(dHBP) \cite{Zhou2011}, tried to improve upon BPFA results with  imposing a prior belief that patches with similar features are likely to be manifested in terms of similar dictionaries. The similarity in this method is defined to be the spatial distance between the location of the patches.

The above reported  works are all concerned with reconstruction of static MRI image slices and not sequences of MRI images as in volumetric (3D) or dynamic MR imaging. Currently these applications, such as real-time cardiac MRI (rtCMR) or functional MRI
(fMRI),   are only possible with a compromise on the
achievable spatial/temporal resolution. High spatial resolution is needed for visualization of fine details or
structures that have diagnostic importance. Simultaneously, the MR image
sequence needs to have high temporal resolution to be able to depict changes
over time due to motion or intensity variation \cite{Jung2009,Zonoobi2014}.   Thus, reconstruction  methods specifically adapted for dealing with dynamic sequences of MRI images, can greatly benefit these applications.


However, not much has been done in the field of DL to deal with sequences of MR images.  Only recently \cite{Wang2014} and \cite{Caballero2014}  extended parametric DL   to  dynamic MRI   by jointly reconstructing the entire sequence and treating it as higher dimensional data.  Through the use of patches, extracted  along both the spatial and temporal directions,  a single spatio-temporal 3D dictionary is trained to  encode  the whole dataset. Sparsity is additionally enforced on the temporal gradient domain as an additional sparsifying transform. Both of these methods fall into category of  \emph{non-causal} (or batch-based)  methods. In non-causal approaches, the entire $T$ frames needs to be acquired before carrying out the reconstruction, which takes advantage of the temporal sparsity.   The main limitation of such methods is their  computational complexity and memory requirement, which  for a $T$-frame acquisition, is roughly  $T^2$ times and $T$ times of that of causal methods, respectively. For example in DLTG method \cite{Caballero2014},  the reconstruction time is reported to be about 6.6 hours  for a sequence of 30 MR images.  The high computational complexity  of DL process  is often avoided by either  learning the dictionary {\it offline} form a training set or using
 only partial patches  to train a global dictionary. Therefore, these dictionaries may fail to sparsely represent patches that are not involved in training \cite{Qu2012}.

 Another limitation of such methods is that online  reconstruction of images  is not possible \cite{Zonoobi2014}.
 Causal approaches  can recover the current frame as soon as its MR data gets acquired, and their memory and computational  demand is much lower than that of non-causal (batch) methods. They are especially  suitable for real-time reconstruction of MRI images. To the best of our knowledge, no DL based method has been proposed for causal reconstruction of MRI sequences.

 In this paper, we propose a  novel DL-based algorithm for online reconstruction of sequence of MR images from highly undersampled $k$-space with the following features:
 \begin{itemize}
\item The algorithm consists of both  patch-based (local) and global sparsity terms. This is based on the observation that each of global and patch-based sparsity  has its own advantages and short-comings (i.e.  local image sparsity do not take into account additional image-level constraints and vise versa).

\item
To avoid the high computational complexity of the DL stage, imposed by the number of patches, and also to train more structured dictionaries,   {\it group patching} is employed to classify the patches based on their similarities. The grouping is done once at the initialization step, using a guide image.


 \item  A modified dHBP is utilized as the prior for the dictionary learning process.  Number of dictionary elements and their relative importance is inferred non-parametrically. In addition, the  model uses patch similarities and spatial closeness to encourage sharing of information within image subregions.

 \item
The method is specially adopted for  reconstruction of dynamic MR images.  To this end,  some useful prior information are extracted from the reconstructed image of the previous time instant.
This priori information guides the reconstruction,  through the  incorporation into  the global sparsity term and also through the initialization stage of the  {\it in situ} DL process, in which dictionaries are learnt   a priori from a fully sampled reference image(s) and then  propagated and updated along the temporal dimension to include new features in the current frame.

\end{itemize}

The rest of the paper is organized as follows. This section ends  with a description of the notations used.
Section \ref{sec:method} presents the  details of  our proposed algorithm as applied to sequences of MRI  images. We present and analyze our experimental results in section \ref{sec:exp} before providing  concluding remarks in section \ref{sec:con}.

{{\bf Notations:} Throughout the paper,  matrices are denoted by capital boldface letters (e.g.  ${\bf D}, {\bf X} $), vectors are denoted by boldface small letters  and we use the notation ${\bf X}_{\mathcal S}$ to denote the sub-matrix containing elements of ${\bf X}$ with indices belonging to set ${\mathcal S}$.  ${\bf X}_{i,j}$  denotes the $(i,j)^{th}$ element of ${\bf X}$. Scalars are shown by regular letters (e.g. $ N, L, n, m, k, r$) and linear maps are denoted by bold calligraphic  uppercase letters ($\R, \A, \P$). Superscript $(t)$ added to a matrix refers to that of time $t$.

\section{Dictionary learning for sequences of MR images}
\label{sec:method}


\begin{figure}[h]
\centering
{\includegraphics[trim=0mm  0mm 0mm 0mm, clip=true, width=0.5\textwidth]{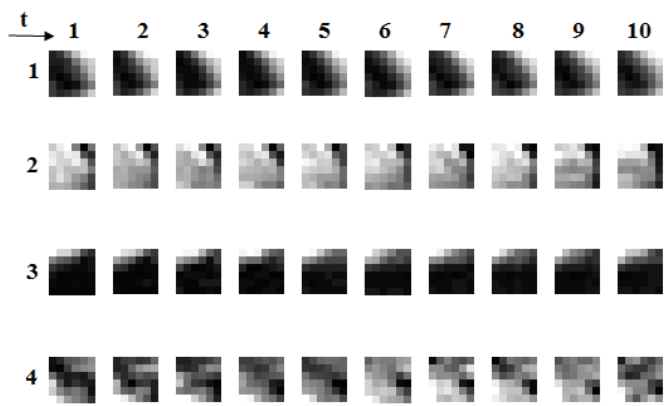}}
\caption{Example of some random patches displayed as sequences in time. }
\label{fig:patchSimilarity}
\end{figure}

The problem can be posed as follows: let $\xt  \in \R^{\sqrt n\times \sqrt n}$ be the slowly time-varying MRI image of interest at epoch $t$ which is   known to have an {\it approximately} sparse representation in some transform domains such as Daubechies wavelets \cite{APSIPA,Zonoobi2013}.  Let $ {\mathcal S}^{(t)}:=\{ (i,j) : (\W \xt)|_{i,j} > \hat\epsilon \} $ denote its support in Wavelet domain. It is assumed that for the first time frame a high sampled image is available (${\bf x}^{(1)}$), from thereafter at each time instant, the under-sampled $k$-space acquisition can be expressed by a linear system given as:  $ {\bf y}^{(t)}= \A (\xt) +{\bf w}$. ${\yt}$ is the observation vector and ${\bf w}$  is the measurement noise with finite energy (i.e. $\|{\bf w}\|_2\le \tilde{\epsilon}$), which can be modeled as a complex Gaussian noise.
We define $\P$ as the patch decomposition operator which extracts  vectorized patches of size $\sqrt {L} \times \sqrt L $ from the image. The $i^{th}$ vectorized patch $\pt_i $ of size $ L$ is expressed as $\pt_i =\P_i\xt$  for $i = 1,...,n$.  These patches overlap with a shift of one pixel and wrap around the image at the boundaries.  Figure \ref{fig:patchSimilarity} shows some random patches of size 6$\times$6 extracted from a sequence of Cardio MR  images
evolving with time. It is not surprising that these patches continue to be closely related to each other as they evolve over time. This property has been used extensively  in  other batch-based dynamic MRI  reconstruction methods by enforcing  the temporal gradient to be sparse \cite{Wang2014,Caballero2014}. We also utilize this property but the key difference is that our method is causal, thus we only assume that at each time epoch $t$, only the reconstructed image of the previous time instant (or a fully sampled reference image) is available.

Each  patch $\p_i$ of image $\xt$, is  assumed to be represented as
a linear combination of a sparse set of atoms from a
dictionary ($\Dt$). The goal  in the simplest setting is to learn $\Dt$ and in so doing infer $\xt$ by solving:

\begin{align}
\label{Eq_1}
\mathop {\min_{\xt,\Dt} }\sum_i & \frac{\gamma_\epsilon}{2} \| \P_i \xt-\Dt\alpha_i \|_{2}  + f(\alpha_i) \nonumber\\ \; \;  &{\rm such\; \;  that} \;  \;  \| \A \xt- { \yt}\|_2 \le \epsilon
\end{align}

where $ f(\alpha_i)$ enforces sparsity on $\alpha$.

Reformulating the above constrained optimization problem using Lagrangian multipliers we get:

\begin{equation}
\label{Eq_2}
\mathop {\min_{\xt,\Dt} }\sum_i  \frac{\gamma_\epsilon}{2}\| \P_i \xt-\Dt \alpha_i \|_{2}  + f(\alpha_i) + \frac{\lambda}{2}  \| \A\xt- { \y}^{(t)}\|_2
\end{equation}
where the weight $\lambda$  depends on the standard deviation  of the measurement noise.
The first and second term in (\ref{Eq_2}) represents the quality of the sparse approximations of the image patches with respect to the trained dictionary while the third one enforces data consistency  in $k$-space.

 The resulting optimization is however  non-convex and of combinatorial nature  and thus one can only hope to reach a local minimum at best \cite{Rubinstein2010}. Different methods have been proposed to solve the above optimization,  however they are all burdened by a heavy computational complexity of the DL process which grows rapidly with the number of patches.  Reducing the number of patches is possible through either increasing each patch size ($L$) or using non-overlapping patches, however it is known that these approaches degrade the performance of  DL-based  methods (see Figure \ref{fig:Patch Size Sensitivity})\cite{Rubinstein2010}.  As an alternative, we make use of the pattern redundancy in the images, which is the root of self-similarity property \cite{Qu2013} and propose to perform  {\it patch grouping} before the DL process. This is discussed in following section.

\subsection{Patch Grouping}

The idea is that instead of using all the patches to train a single global dictionary, we classify the patches based on their similarities, so that each group only  contains patches that are closely related to each other and then train different dictionaries for each group.
We found that benefit of this approach is two-fold.
Firstly, the number of patches is reduced for each DL process and since
the learning stage is separate for each class, DL can be performed concurrently
for all the classes which significantly speeds up the computation.
Secondly, since the patches in each class are closely related, it is anticipated
that the trained dictionaries become more structured and tailored to that specific group (see Figure \ref{fig:dic}).

To this end we aim to partition  patches into $N_g$ groups so that each patch belongs to only one group.
We also define similarity as the $L_2$ norm distance between  patches' intensity, where a smaller distance corresponds to  more similarity.
To this end, we have used {\it k-mean} \cite{Ahmad2007} as a fast and efficient heuristic algorithm, to perform the grouping.

Since the $k$-space data is undersampled, there are no available ground truth images
to learn the similarity.   In our proposed method, the grouping is guided based on the assumption
that  ${\bf x^{(1)}}$ (or a training image)  is available and that the patches are changing slowly with time, we propose to  use this guide image for the grouping.  As discussed in greater detail in section \ref{sec:exp}, this is found to be more efficient than grouping the patches based on  the zero-filled reconstructed image. Moreover, this grouping can  be even done off-line before the acquisition process.
 If ${\bf G_j}$ for  $j=1,...,N_g$ stores the index of patches belonging to the $j^{th}$ group,  (\ref{Eq_2}) can be rewritten as:
\begin{equation}
\label{Eq_3}
\mathop {\min_{\xt,\Dt} }\sum_{j=1}^{N_g}\sum_{i\in G_j}  \frac{\gamma_\epsilon}{2} \| \P_i \xt-\D_j^{(t)} \alpha_i \|_{2}  + f(\alpha_i) + \frac{\lambda}{2}  \| \A\xt- { \y}^{(t)}\|_2
\end{equation}
where  $\D_j   \in \R^{m_j \times  k_j}$  is the dictionary learnt for each group $j$.  Figure \ref{fig:patchGrouping} shows an example of how patches are grouped for a Cardio image.

\subsection{Global sparsity and Priori knowledge}

A main limitation of the DL methods that are only based on patch-level (local) sparsity is that they do not take image-level constraints such as
smoothness into account. Our aim is to first add a global sparsity term to our problem formulation and secondly, to make use of any available priori-knowledge by incorporating it into the reconstruction process.
To this end,  we combine a global sparsity term, as the regularization of the reconstruction, by enforcing the image of interest to be sparse in the Wavelets domain.  Moreover, we  incorporate the priori-knowledge which is available from a reference image ($\x^{(r)}$) to guide the reconstruction of the current image of interest ($\xt$). Inspired by recent works  \cite{ModCS2011,Zonoobi2014} that exploited the Wavelet sparsity of each  frame of an image sequence with respect to a reference image, we
 also extend the use of  such a-priori into to DL process.
 For the global sparsity, we use the support of the reference frame ($ {\mathcal S}^{(r)}$) as a  close estimate to the support of the signal of interest ($\xt$) and then use this estimate to reconstruct $\xt$ by finding a signal which satisfies the observations and is sparsest outside ${\mathcal S}^{(r)}$ in the Wavelets domain. This is equivalent to solving the following optimization problem:
\begin{align}
\label{Eq_4}
\mathop {\min_{\xt,\D_j^{(t)}} } \lambda_g  \|  \W_{  \hat {  \mathcal  S}^{(r)}} \xt \|_1
& +  \sum_{j=1}^{N_g} \sum_{i\in G_j} \frac{\gamma_\epsilon}{2} \| \P_i \xt  - \D_j^{(t)} \alpha_i \|_2^2  \nonumber \\ &+   \| \alpha_i \|_0  +      \frac{\lambda}{2} \| \A \xt- \yt\|_2^2
\end{align}
where ${\bf x}^{(r)}$ is the reference image which can either be ${\bf x}^{(t-1)}$ or ${\bf x}^{(1)}$  and $\W_{  \hat {  \mathcal  S}^{(r)}} \x := \{ \W_{ i,j} \x: (i,j)\notin   {\mathcal S}^{(r)}\}$.
As for the incorporation of the priori-knowledge into the DL process, this is done through initialization stage of the optimization and is explained in further detail  in section \ref{sec:7b}.

\subsection{Optimization algorithm}
To solve the optimization problem (\ref{Eq_4}) we use a simple and efficient method based on the {\it Alternating Direction Method of Multipliers }(ADMM) \cite{Boyd2011,Xinghao2013}.  The idea is to use an additional augmented Lagrangian term to split the objective function into different conditionally independent and separable components  and then solve each part sequentially  until it converges. To this end  a new  variable ($\bf v$) is introduced such that $\bf v = \W_{  \hat {  \mathcal  S}^{(r)}} \xt$. Then (\ref{Eq_4}) can be re-written as
\begin{align}
\label{Eq_5}
&\mathop {\min_{\xt,\D_j^{(t)} } } \lambda_g  \| \v\|_1
				+  \sum_{j=1}^{N_g} \sum_{i\in G_j} \frac{\gamma_\epsilon}{2} \| \P_i \xt- \D_j^{(t)} \alpha_i \|_2^2
			\nonumber \\&+ \| \alpha_i \|_0 +     \frac{\lambda}{2} \| \A \xt- \yt\|_2^2 				 \; \;  {\rm such\; \;  that} \;  \;  \bf v - \W_{  \hat {  \mathcal  S}^{(r)}} \x=0
\end{align}

The augmented Lagrangian form of (\ref{Eq_5}) is:
\begin{align}
\label{Eq_6}
&\mathop {\min_{\xt,\Dt,\v,\u} } \lambda_g  \|  \v \|_2 + \u^T ( \W_{  \hat {  \mathcal  S}^{(r)}} \xt- \v ) +  \frac{\rho}{2} \| \W \xt- \v  \|_2^2   	\nonumber \\
			&	+  \sum_{j=1}^{N_g} \sum_{i\in G_j} \frac{\gamma_\epsilon}{2} \| \P_i \xt- \D_j^{(t)} \alpha_i \|_2^2
			+ \| \alpha_i \|_0 +     \frac{\lambda}{2} \| \A \xt- \yt\|_2^2
\end{align}
where $\u$ is the augmented Lagrangian dual variable and we add a scalar $\rho$ before $\u$ to make things simple without affecting the results. Now, (\ref{Eq_6}) can be decomposed into three separate minimization problems as:
\begin{subequations}
\begin{align}
&    \tilde{\bf v} = \arg \min_v  \lambda_g \|{\bf v} \|_1 + \frac{\rho}{2} \| \W_{  \hat {  \mathcal  S}^{(r)}} \xt-{\bf v}  \|_2^2 + u^T ( \W_{  \hat {  \mathcal  S}^{(r)}} \xt- {\bf v} ) 	\label{eqn:7a}\\
& (\tilde \D_j^{(t)}, \tilde \alpha_i) = \arg \min_{\alpha_i,\D_j^{(t)}}  \sum_{j=1}^{N_g} \sum_{i\in G_j} \frac{\gamma_\epsilon}{2} \| \P_i \xt- \D^{(t)}_j \alpha_i \|_2^2  + f(\alpha_i) 	\label{eqn:7b}\\
& \tilde {\bf x}^{(t)}  = \arg \min_{\x} \frac{\rho}{2} \| \W \xt- \tilde {\bf v}  \|_2^2 + \u^T ( \W_{  \hat {  \mathcal  S}^{(r)}} \xt- \tilde  {\bf v} )  \nonumber\\
& +  \sum_{j=1}^{N_g} \sum_{i\in G_j}  \frac{\gamma_\epsilon}{2} \| \P_i \xt- \tilde {\bf D}_j^{(t)} \alpha_i \|_2^2  +   \frac{\lambda}{2} \| \A \xt- \yt\|_2^2  \label{eqn:7c}\\
&  \tilde \u= \u+ \W_{  \hat {  \mathcal  S}^{(r)}}  \xt-  \tilde {\bf v}  \label{eqn:7d}
\end{align}
\end{subequations}

 where first in (7a) we fix $\xt$ and $\D_j^{(t)}$ and solve for $\v$. Then  using the updated value of $\v$ ($  \tilde{\bf v}$) we learn the dictionaries from (7b). With the obtained  $\tilde\v$ and $\tilde \D^{(t)}$, we then reconstruct $\xt$ such that it is consistent with the  $k-$space measurements  by solving the minimization of (7c) over $\x$. Finally we update the dual variable $\u$. These steps are repeated until convergence (see Algorithm 2). The details of solving the above minimization problems are as follows:

\paragraph{Solving (\ref{eqn:7a})}

Optimization  (\ref{eqn:7a})  is known to be equivalent to  a soft-shrinkage problem \cite{Boyd2011} with a closed form solution of:
\begin{align}
\label{Eq_8}
\quad \tilde {\bf v} = {\bf \mathbb{S}}_{\frac{\lambda_g}{\rho}}( \W_{  \hat {  \mathcal  S}^{(r)}}  (\tilde \x)+\tilde u)
\end{align}
where  ${\bf \mathbb{S}}_{\kappa}(x)=
\left (
1-\frac{\kappa}{\|\x\|_2}
\right ) $  is the soft-shrinkage operator \cite{Goldstein2008}.

\paragraph{Solving (\ref{eqn:7b})}

\label{sec:7b}



To solve the non-convex objective function (7b),  we use a   nonparametric Bayesian-based method. Here the goal is to compute  ${\bf D_j^{(t)}} \in  \R^{m_j*K_j}$ for $j=1,...,N_g$ and the corresponding sparse coefficients $\alpha$ using a stochastic optimization method.  To achieve this, a beta process is used as a prior density and consequently, the posterior density of the dictionaries and the parameters of the model is inferred using the Gibbs sampling method \cite{Damlen1999}.
Let us define  $\alpha$ to be the elementwise product of a Gaussian vector ($\s$) with a binary vector $\z$. Thus, for each patch ${\bf x^{(t)}_i}$ belonging to group $j$, i.e. $i \in G_j$ we have $${\bf x^{(t)}_i} =  {\bf D_j^{(t)}}(\s_i \odot  \z_i)  + \epsilon_i$$
where  $\odot$  represents the  Hadamard product, $\s_i = [s_{i,1},\cdots,s_{i,k_j}]^T $,
$\z_i = [z_{i,1}, \cdots,z_{i,k_j}]^T$, $s_{i,k} \in \R$ and $z_{i,k} \in\{0, 1\}$ indicates whether  for the patch $i$, the $k^{th}$ dictionary element is \emph{active} or not,  $\epsilon_i$ is also the residual error. For each group $j$, the dictionary size ($k_j$) is initially  set  to be large, and its actual  value is inferred  using Gibbs sampling.

Our goal is to incorporate  two important features into our Bayesian nonparametric method: First, based on an intuition that similar patches in an image are consist of similar dictionary atoms,  we want to encourage  patches close to each other  in one group, in terms of a defined similarity measure, to  utilize similar values of $\z$.
Second, based on the assumption that a reference image is available, and also that the image of each frame is closely related to the one of the previous time instant, we want to design a dynamic method such that  it uses the reference image as an initialization guide, then it propagates/updates the dictionaries along the temporal direction to include new features of the current frame.

To this end, we have adopted the {\it  dependent hierarchical beta process} (DHBP) \cite{Zhou2011} to find the dictionaries for each group of patches. In addition, we consider the similarity of patches in each frame and the information of first frame in the model construction.
To achieve a sparse representation for the coefficients $\z_i$, a beta process is assigned to it in the initialization step. Moreover, the distance  between the patches of each group is also  considered in the initialization step.

We define a  kernel that shows the similarity of two patches and takes both the spatial location and intensity values of both patches into account as:

\begin{align*}
K({\bf p}^{(t)}_i,{\bf p}^{(t')}_j) =  \left\{\begin{array}{lcl}0 \qquad & if& \nexists k :  {\bf p}^{(t')}_j,{\bf p}^{(t)}_i  \in G_k  \\0 & if &   \| i- j \|_2 \le R_1 \\ e^{ (-\| {\bf p}^{(t)}_i - {\bf p}^{(t')}_j \| / \sigma) }
&  & else \end{array}\right.
\end{align*}

 where  $\|\p^{(t)}_i - \p^{(t')}_j||$ represents the intensity difference of two patches and $\sigma$ is the kernel width. Note that $K$ assumes a maximum value of 1 in the case where two patches are identical and diminishes with increasing distance between the pixel values of  ${\bf p}_i$ and ${\bf p}_j$. It becomes 0 if  two patches do not belong to a same group or if they are not located in the spatial vicinity (neighborhood) of each other.
We then define a matrix ${\bf A}$ such that each of its  rows  sums to one as:
\begin{align*}
{\bf A} _{i,j} = K(\p^{t}_i,\p^{t'}_j)  \big/  \sum_{j' = 1}^{N_j} K(\p^{t}_i,\p^{t'}_{j'})
\end{align*}

Matrix  ${\bf A}$, which manifest the interrelation of all the patches, is then used in
an analogous manner to the covariates  matrix in \cite{Zhou2011},  to impose a  priori-belief such   that similar patches  are more likely   to employ similar dictionary settings.

 It should be noted that at the initialization step of our algorithm, the kernel represents the similarity of patches for the current and first frames (reference frame) $(t' = 1)$, while in the other iterations, it shows the similarity of patches with each other on the same frame $(t' = t)$.


The parameters of the model are initialized in the hierarchical framework as described in algorithm 1.
For the first frame, the columns of $\D_j$, $\epsilon_i$ and $\s_i$ is initialized with a Gaussian prior and a beta-Bernoulli process is placed as a prior for
$z_i$. Moreover, a gamma hyper-priors is considered for the parameters of $\s_i$ and $\epsilon_i$ ($\gamma_s$ and $\gamma_e$). For the subsequent frames, the parameters $\D_j$, $\epsilon_i$ and $\s_i$ are initialized using the posterior density of the first frame. Moreover, $z_i$ is a Bernoulli process of the weighted summation of beta processes. The weight is defined using the matrix $A(p^{(t)}_i,p^{(1)}_j)$. The parameters of the DNBG method which should be learned using Markov Chain Monte Carlo (MCMC) algorithm are  $(D_j^{(t)},z_i,s_i,\epsilon_i,\pi,\eta,\gamma_s,\gamma_e)$.  The posterior density of these paprmeters is calculated in the appendix 1.

\begin{figure*}[htp!]
\centering
{\includegraphics[trim=10mm  10mm 10mm 0mm, clip=true, width=0.7\textwidth]{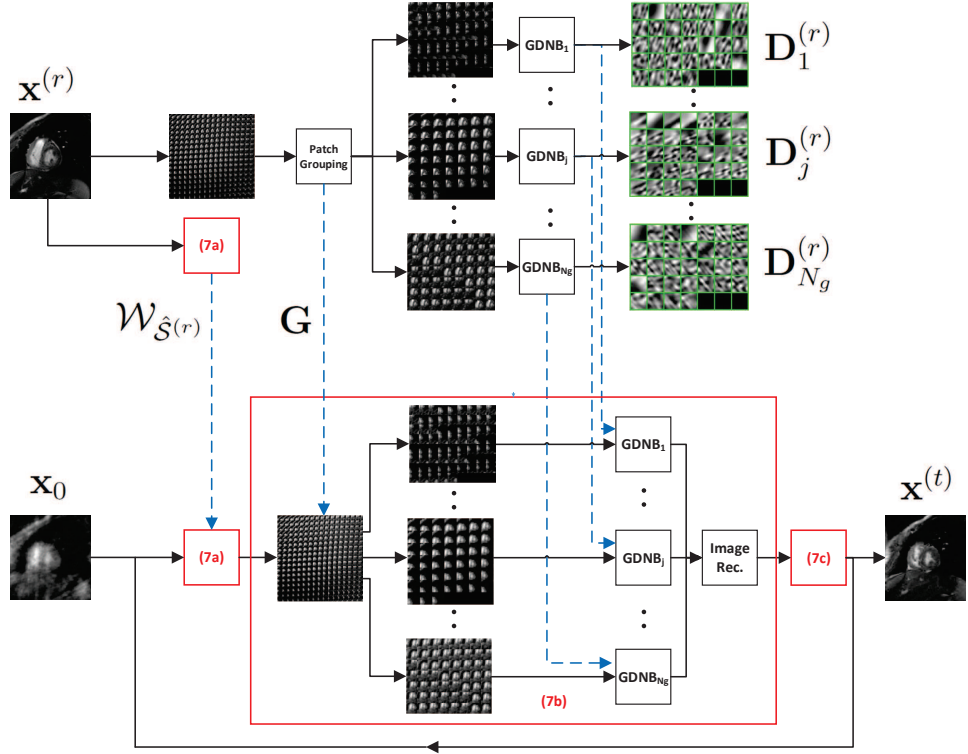}}
\caption{Block Diagram of the proposed method.}
\label{fig:patchGrouping}
\end{figure*}

\begin{algorithm}[h]
\begin{enumerate}
\item [(a)] {\bf Construct $N_g$ number of dictionaries as ${\bf D_j^{(t)}}= [{\bf d_{j1}^{(t)}},....,{\bf d_{jk}^{(t)}}]$: \\${\bf d_{ji}^{(1)}} \sim \N(0,L^{-1}I_L)$,  for $i=1,...,K$ $j=1,...,N_g$ \\ ${\bf D_j^{(t)}} ={\bf D_j^{(1)}}$, for $j=1,...,N_g$ }
 \item[(b)] {\bf   Draw a probability $\pi_{ik} \in [0, 1]$ for each element of $\D$:}
 \\ $\pi_{ik} = \sum_{l \in Q_i}a_{il}\pi^*_{lk}$
 \\$\pi^*_{lk} \sim Beta (c_1\eta_k, c_1(1-\eta_k))$
 \\$\eta_k \sim Beta(c_0\eta_0, c_0(1-\eta_0))$
  \item[(c)] {\bf  Draw precision values for noise and each weight \\ $\gamma_{\epsilon}\sim Gamma(g_0,h_0)$, \;\;\; $\gamma_{s,k}\sim Gamma(e_0,f_0)$}
  \item[(d)] {\bf  For the $i^{th}$ patch in $\xt$:\\}
  (i)Draw the vector $s_i \sim \N(0,diag(\gamma^{-1}_{s,k}))$.\\
 (ii)Draw the binary vector $z_i$ with elements \\
 $z_{ik} \sim Bernoulli(\pi_{ik}).$\\
  (iii)Define $\alpha_i=s_i \circ z_i $by an element-wise product. \\
   (iv)Construct the patch  $\P_i( \x)= \D \alpha_i+\epsilon_i $with noise. \\
   $\epsilon_i \sim \N(0,\gamma_{\epsilon}^{-1}I_L).$
     \item[(e)]  {\bf  Construct the image $\x$ as the average of all patches.}\\
    \end{enumerate}
\caption{Priori DNBG initialization }
\label{alg:BPFA}
\end{algorithm}

\begin{algorithm}[h]
\textbf{\texttt{Input}:}{ $\yt$}, $\x^{(1)}$\\
\textbf{\texttt{Output}:}  ${\bf x}^{(t)}$
\begin{enumerate}
\item [(a)] {\bf Initialization: dictionary variables using Algorithm1, $\x_0 = \A\yt$ , $\u= 0$.}\\
 \item[(b)] {\bf  While {\ not  converged } do}\\
  (i) Solve (7a) sub-optimization using (\ref{Eq_8})\\
  (ii) Update (7b) sub-problem by Gibbs sampling dictionary variables.\\
  (iii) Solve (7c) sub-optimization using (\ref{Eq_9} ) in Fourier domain, followed by inverse transform\\
  (iv) Update the dual vector $u$ using (7d).
    \end{enumerate}
\caption{DNBG Algorithm for reconstruction of $\xt$ from $\yt$.}
\label{alg:BPFA-priori}
\end{algorithm}

\paragraph{Solving (\ref{eqn:7c})}
Problem (\ref{eqn:7c}) is a  least squares problem with an analytical solution given as:

\begin{align}
\label{Eq_9}
 & \left (\gamma_\epsilon  \sum_{j=1}^{N_g} \sum_{i\in G_j}  \P_i^T\P_i+ \lambda \A^T\A+\rho  \W^T_{  \hat {  \mathcal  S}^{(r)} }  \W_{  \hat {  \mathcal  S}^{(r)} }  \right )^{-1} \tilde {\bf x}^{(t)}  = \nonumber \\
& \left (\gamma_\epsilon  \sum_{j=1}^{N_g} \sum_{i\in G_j}  \P_i^T  \D_j^{(t)} \alpha_i +\lambda \A^T \yt +\rho \W^T_{  \hat {  \mathcal  S}^{(r)} } (\tilde \v- \tilde \u) \right )
\end{align}

Note that term $\sum_{j=1}^{N_g} \sum_{i\in G_j}  \P_i^T  \Dt \alpha_i $
represents the reconstructed image  obtained by averaging the contributions of the various patches. We observe that inverting the left  matrix is computationally prohibitive. However, it is known from \cite{Goldstein2008} that a simplification can be obtained by transforming from image space to Fourier  space. The overview of our proposed method, called {\it Dependent Nonparametric Bayesian Group Dictionary Learning} (DNBG) is shown in Figure \ref{fig:patchGrouping} and also Algorithm 2.



\section{Experimental Results}
\label{sec:exp}




The proposed DNBG method  was tested on 3 sequences of MRI images (see details in Table \ref{tab:sepcs}\footnote{Cardiac data set was provided by the Department of Diagnostic Imaging of the Hospital for Sick Children in Toronto, Canada.  These images were scanned with a GE Genesis Signa MR scanner using the FIESTA scan protocol. The single slice brain perfusion MRI data set was obtained from amulti slice 2-D dynamic contrast enhanced (DCE) patient scan at the University of Rochester \cite{Lingala2013} \kkk and the thorax data set  were acquired at the National University Hospital, Singapore. } )   some of which are shown in Figure \ref{fig:seq}.


\begin{figure}[h]
\centering
{\includegraphics[trim=60mm  15mm 40mm 0mm, width=0.45\textwidth]{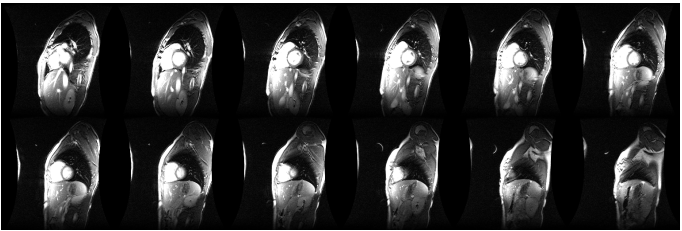}}
\caption{Cardiac MRI sequence image samples. }
\label{fig:seq}
\end{figure}


\begin{table}[h]
\caption{ Specifications of data sets used in the experiments.}
\centering
\setlength{\tabcolsep}{1 pt}
\renewcommand{\arraystretch}{1}
\begin{tabular}{|c||c c c |}
\hline
Data set& Cardiac  &  Brain & Thorax\\ \hline\hline
{\scriptsize Bits allocated}  &8&8&16 \\
{\scriptsize Rows$ \times$ columns  }  & {\scriptsize  256$ \times$ 256 } &{\scriptsize128$ \times$ 128}& {\scriptsize 384$ \times$ 384} \\
{\scriptsize Slices}&20&60&20 \\
{\scriptsize Pixel size (mm)}& {\scriptsize 0.93 $ \times$ 0.93 }&{ --}&{\scriptsize  0.64 $ \times$  0.64} \\
{\scriptsize Inter-slice spacing}&6&--&0.9\\
{\scriptsize Modality}&MRI&MRI&MRA\\ \hline
\end{tabular}
\label{tab:sepcs}
\end{table}

\begin{figure}[h]
\centering
\subfigure
{\includegraphics[trim=20mm 10mm 10mm 1mm, clip=true,width=0.20\textwidth]{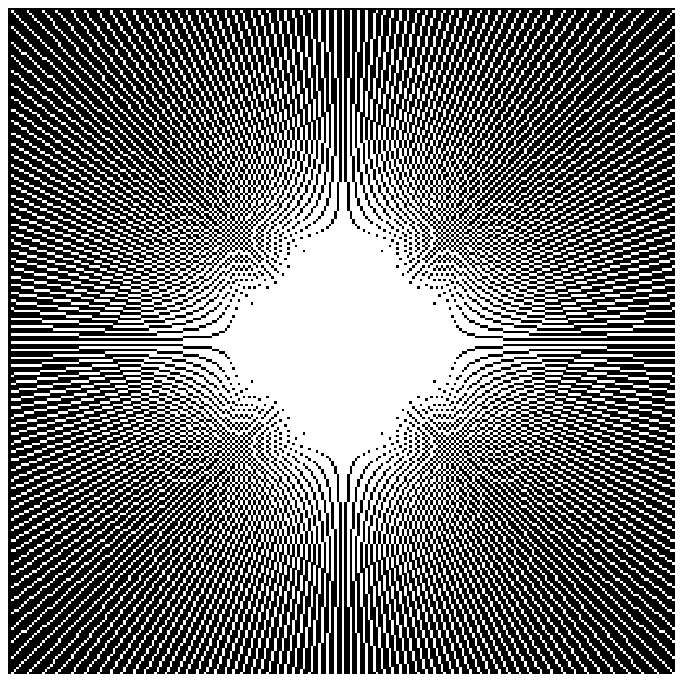}}
\subfigure
{\includegraphics[trim=20mm  10mm 10mm 0mm,clip=true, width=0.20\textwidth]{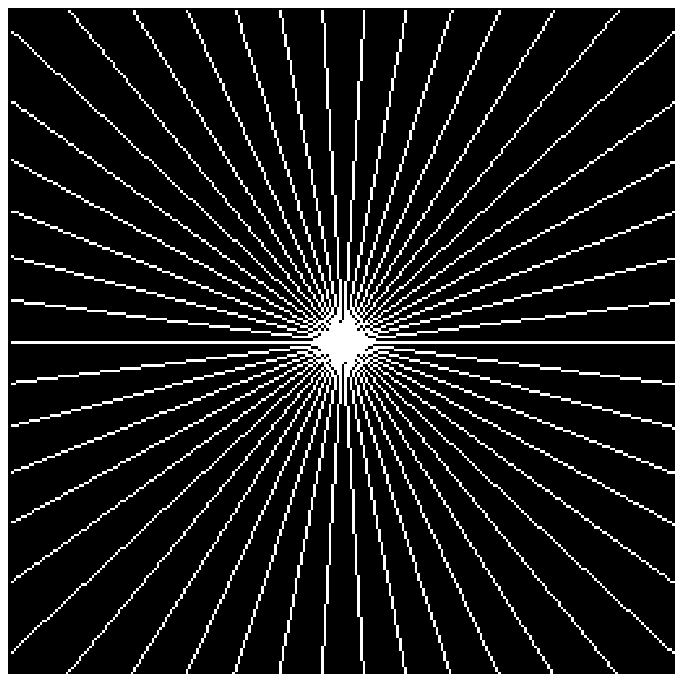}}
\caption{Radial sampling mask for (left) t=1 and (right) subsequent frames. }
\label{fig:mask}
\end{figure}

\paragraph{Sampling} Radial sampling with  uniformly spaced
rays in the $k$-space domain, has been widely used for compressed MR image reconstruction \cite{Wang2014} as it provides an incoherent measurement basis. Figure \ref{fig:mask} shows the sampling masks used in our experiments with two different sampling ratios R, defined as $m/(\sqrt n\times \sqrt n)$.  For the first image frame ($t = 1$) in each sequence, we set R={\large$\frac{2}{5}$ }(Figure \ref{fig:mask}(a))  since no priori knowledge is available while for the successive frames, we consider several subsampling rates (Figure \ref{fig:mask}(b)).

\paragraph{Parameters settings}
Patches are extracted  from the images  in size of 4$\times$4   and the dictionaries' initial size is set to be $\bf K_j$ = 128.  Recall  that the actual number of dictionary elements is inferred  to be a smaller number. Number of groups $N_g$ is also set to 11 and the neighborhood (search) range $R_1$ is assumed to be 13. The choice of these parameters that are specific to the DNBG method is discussed  in section \ref{sec:disscussion}. Moreover, assuming a  noiseless sampling, we set the data consistency parameter $\lambda = 10^{10}$, i.e. we only allow DNBG to fill in the
missing k-space. We also set $\lambda_g = 10$, $\rho = 1000$, $c_0 = 1, c_1 = 1$, $e_0 = f_0 = 1$, $\gamma$  = , $g_0 = 1$, $h_0 = 1$ and $\eta_0 = 1$. We run 100 iterations of the algorithm 2
and present the results of the last iteration.

\paragraph{Performance measures}
To evaluate the quality of the reconstruction,  we use  the Peak signal-to-noise ratio (PSNR) to measure the difference of reconstructed image and the fully sampled image which is assumed to be the ground truth as:$$PSNR(\xt_{Rec})=10\log_{10} {1/}{ {MSE(\xt_{Rec},\xt_{F})}}$$
where, $\xt_{F}$ is the fully sampled image and $\xt_{Rec}$ is the fully sampled image of time $t$.

\subsection{ Reconstruction results}

We compared our algorithm with three other state-of-the art DL-based methods. (a) DLMRI  method \cite{Ravishankar2011}, which is based on K-SVD, (b) BPFA \cite{Huang2014}  which is a nonparametric bayesian dictionary learning method and (c) Blind Compressive Sensing (BCS) \cite{Lingala2013} which is the current  state-of-the-art batch-based method  specifically developed for the reconstruction of  MR sequences \footnote{ \kkk The implementation codes for  DLMRI and BCS  methods  were obtained from \url{http://www.ifp.illinois.edu/~yoram/DLMRI-Lab/DLMRI.html} and
 \url{http://research.engineering.uiowa.edu/cbig/content/software}}.

\begin{figure}[h!]
\centering
\subfigure[Cardiac]
{\includegraphics[trim=8mm  0mm 10mm 5mm, clip=true, width=0.45\textwidth]{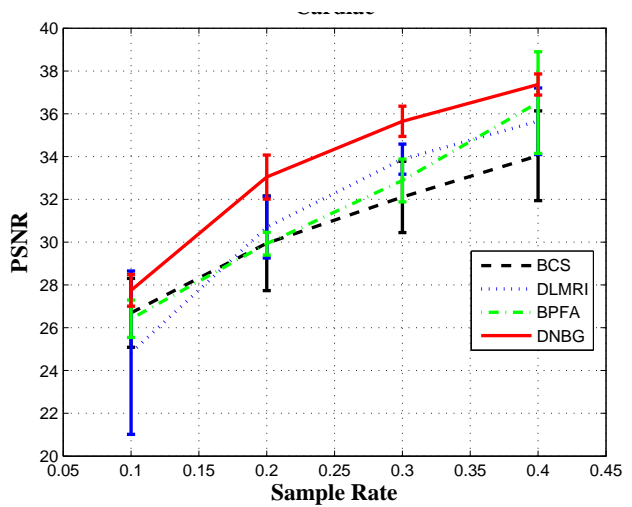}}
\subfigure[Brain]
{\includegraphics[trim=8mm  0mm 10mm 6mm, clip=true, width=0.45\textwidth]{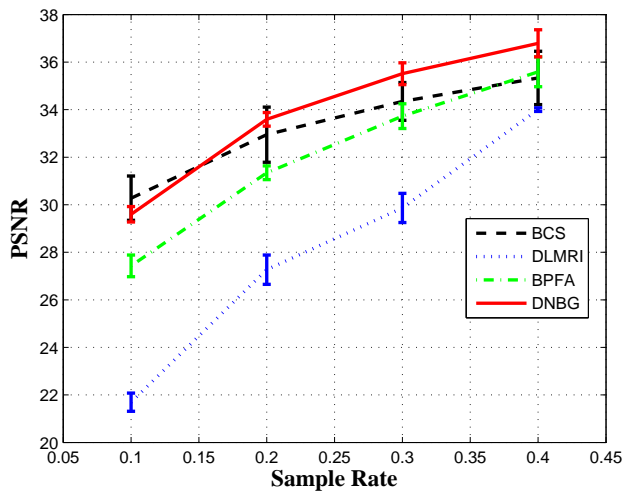}}
 \subfigure[Thorax]
{\includegraphics[trim=8mm  0mm 10mm 6mm, clip=true, width=0.45\textwidth]{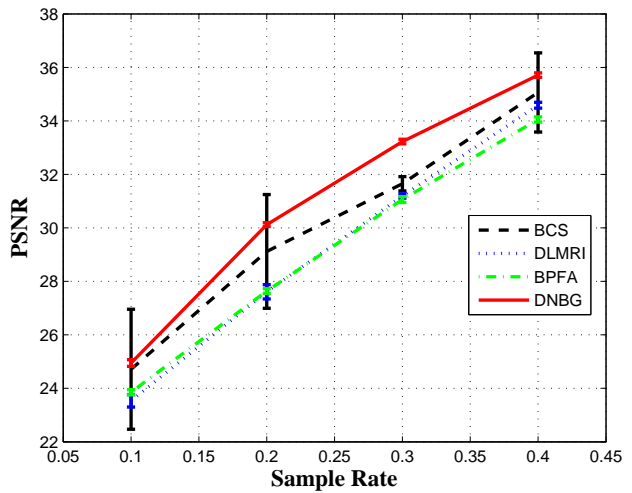}}
\caption{Mean and standard deviation PSNR  of the reconstructed images vs. the sampling rate. }
\label{fig:PSNR}
\end{figure}

Figure \ref{fig:PSNR} shows the reconstruction performance of our  proposed method, compared against other methods, in terms of the PSNR value at different k-space sample rates. For each sample rate the mean and standard deviation of reconstructed result of all  frames has been shown. It can be seen that for all data sets,  DNBG  out-performs  other methods, in terms of both higher mean performance and smaller standard deviation of the reconstruction results. This is more profound for the cardiac data set. Although BPFA and DNBG both utilize  nonparametric bayesian model for the DL, the PSNR of DNBG is significantly higher due to the grouping  and incorporation the dependency of patches. It is also observed that  in some settings BCS tends to achieve  comparable mean PSNR performance to our method. However, its std is consistently  much higher than our method i.e. its performance maybe quite different from one frame to another.
\begin{figure*}[h]
\centering
{\includegraphics[trim=3mm  130mm 0mm 35mm, clip=true, width=0.9\textwidth]{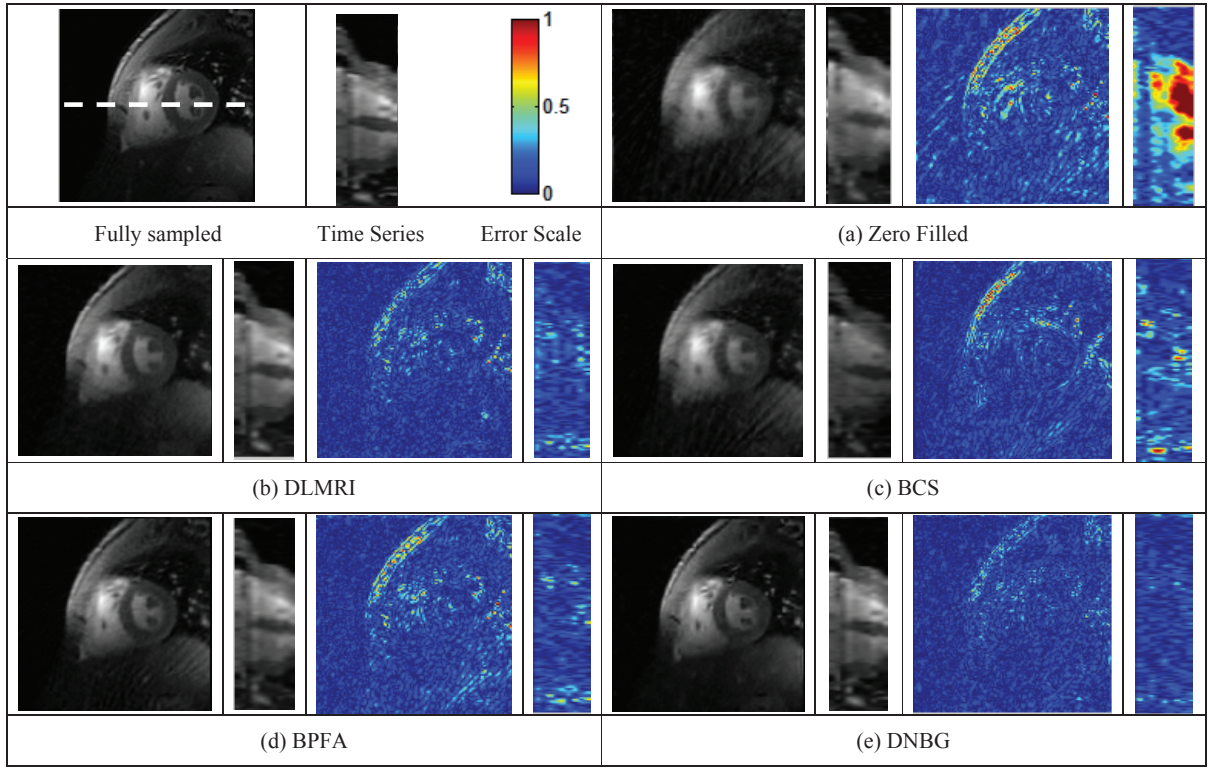}}
\caption{Comparison of the reconstruction results for the Cardiac data set with the sampling  rate of 0.2.}
\label{fig:CompareMethods1}
\end{figure*}

To also compare the visual quality of the reconstructed images, Figures \ref{fig:CompareMethods1}-\ref{fig:CompareMethods3}  show the reconstruction result of our method against others for the cardiac, brain and thorax data sets, respectively. The original frame, together with the original image time profile through the dotted line is shown in the top left cell. For each method the reconstruction image and time profile, together with the error amplified by a factor of 4 is also shown. Note that for the cardiac and thorax data set, what is shown is a ROI in which we have highest variation in time.  From these figures, it is evident that our method achieved superior visual reconstruction quality. It is specially noted that  the reconstruction error of  other methods is mainly concentrated near the boundaries that are changing.


 \begin{figure*}[h]
\centering
{\includegraphics[trim=3mm  130mm 0mm 35mm, clip=true, width=0.9\textwidth]{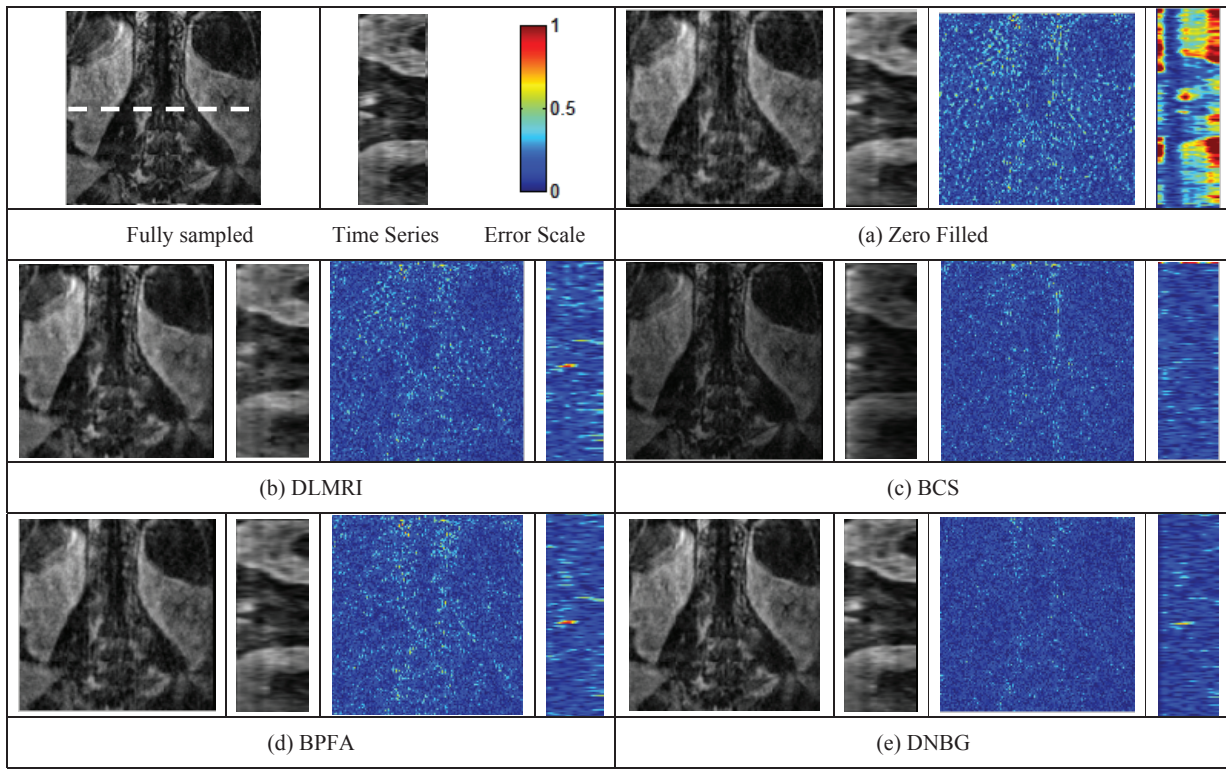}}
\caption{Comparison of the reconstruction results for the Thorax data set with the sampling  rate of 0.2.}
\label{fig:CompareMethods3}
\end{figure*}

\clearpage

  \begin{figure*}[h!]
\centering
{\includegraphics[trim=3mm  130mm 0mm 35mm, clip=true, width=1\textwidth]  {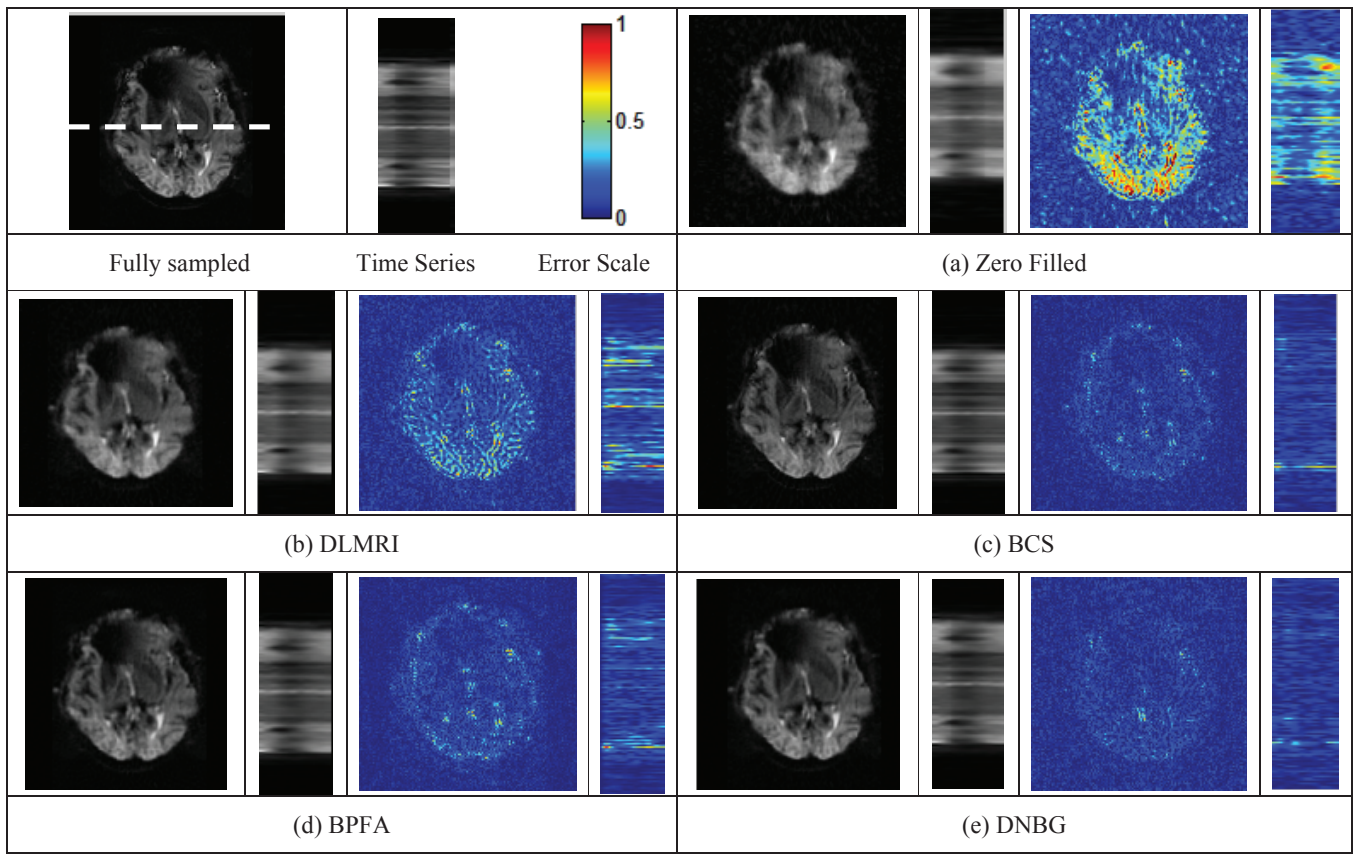}}
\caption{Comparison of the reconstruction results for the Brain data set with the sampling  rate of 0.2.}
\label{fig:CompareMethods2}
\end{figure*}
\clearpage
\subsection{Discussion about the choise of parameters}
\label{sec:disscussion}
The performance of the proposed method depends on the number of patch groups ($N_g$),  patch size ($L$) and neighborhood range ($R_1$).  We now discuss the behavior of our method with respect to changes in these parameters.

 \paragraph{Dependence on  $N_g$}  Figure \ref{fig:Patch Group Number Sensitivity}(a),   shows the run-time of one iteration of our algorithm vs. the  number of groups ($N_g$).  From the figure it can be seen that grouping the patches even into 2 groups can significantly speed-up the computations. It is observed that  up to $N_g=4$ the computation complexity of the DL process  is inversely proportional to the number of groups and thereafter it is almost constant.    On the other hand, Figure \ref{fig:Patch Group Number Sensitivity}(b) shows the mean and variance of the reconstruction error as a function of $N_g$.  It can be seen that  the performance improves as $N_g$ increases and is the highest possible for a range of $N_g=10$ up to $12$ beyond which the error begins to increase.  This observation is inline with Figure \ref{fig:Patch Group Number Sensitivity}(c), which shows the sparsity of $\alpha$, measured using the Gini Index (GI) \cite{Zonoobi2011},   as a function of $N_g$. It can be seen that increasing  group number also increases the sparsity of $\alpha$ (since the GI is closest to 1), which means that smaller number of dictionary atoms are needed  to represent a patch. This could be due to the reason that the trained dictionaries are more tailored to that specific group and are better able to represent a patch belonging to that class.  On the other hand,  increasing $N_g$ beyond 12 decreases $\alpha$'s sparsity significantly.

   \begin{figure}[h]
\centering
\subfigure[]
{\includegraphics[trim=10mm  0mm 15mm 5mm, clip=true,width=0.5\textwidth]{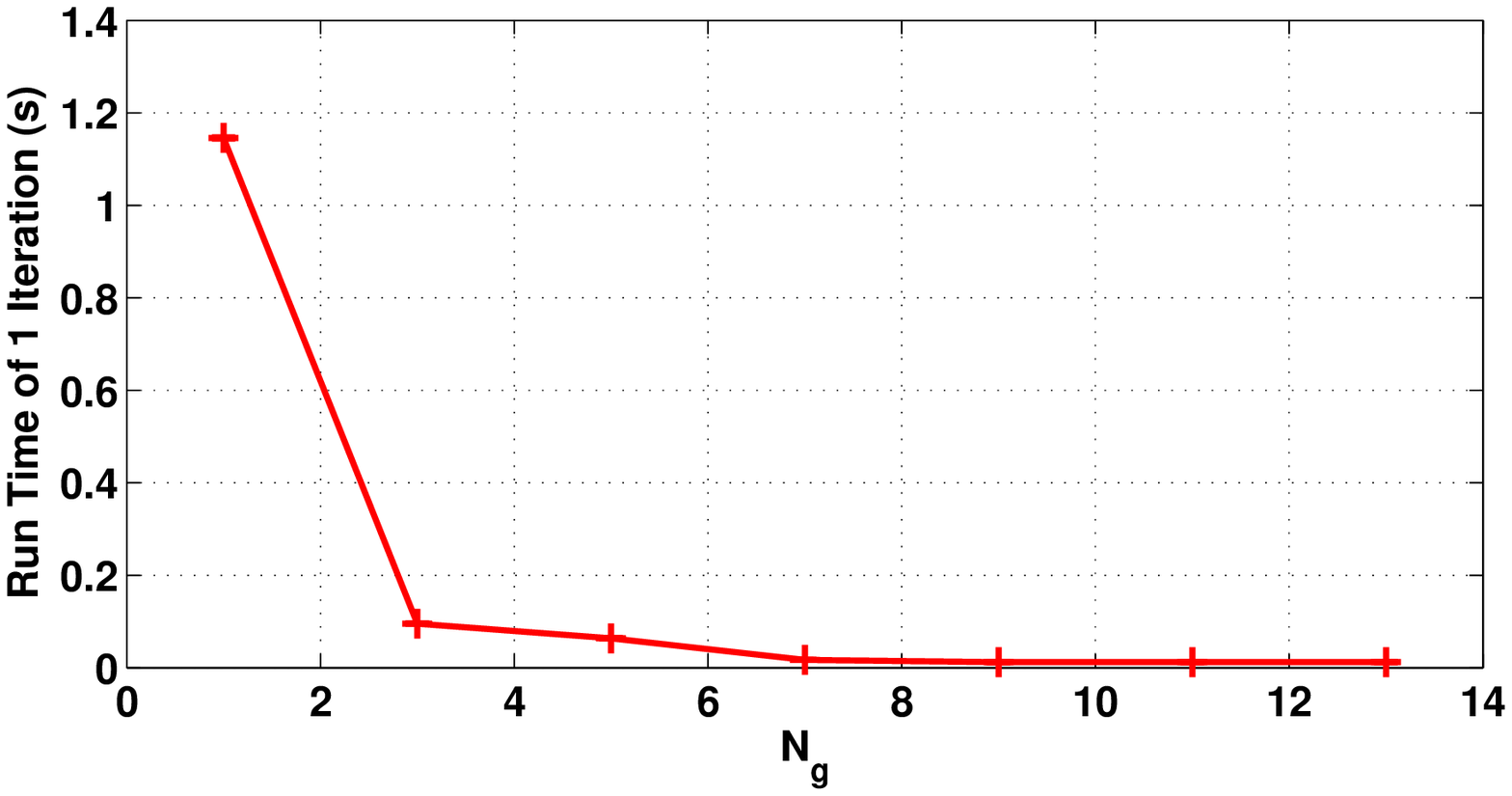}}
\subfigure[]
{\includegraphics[trim=10mm  0mm 15mm 5mm, clip=true,width=0.5\textwidth]{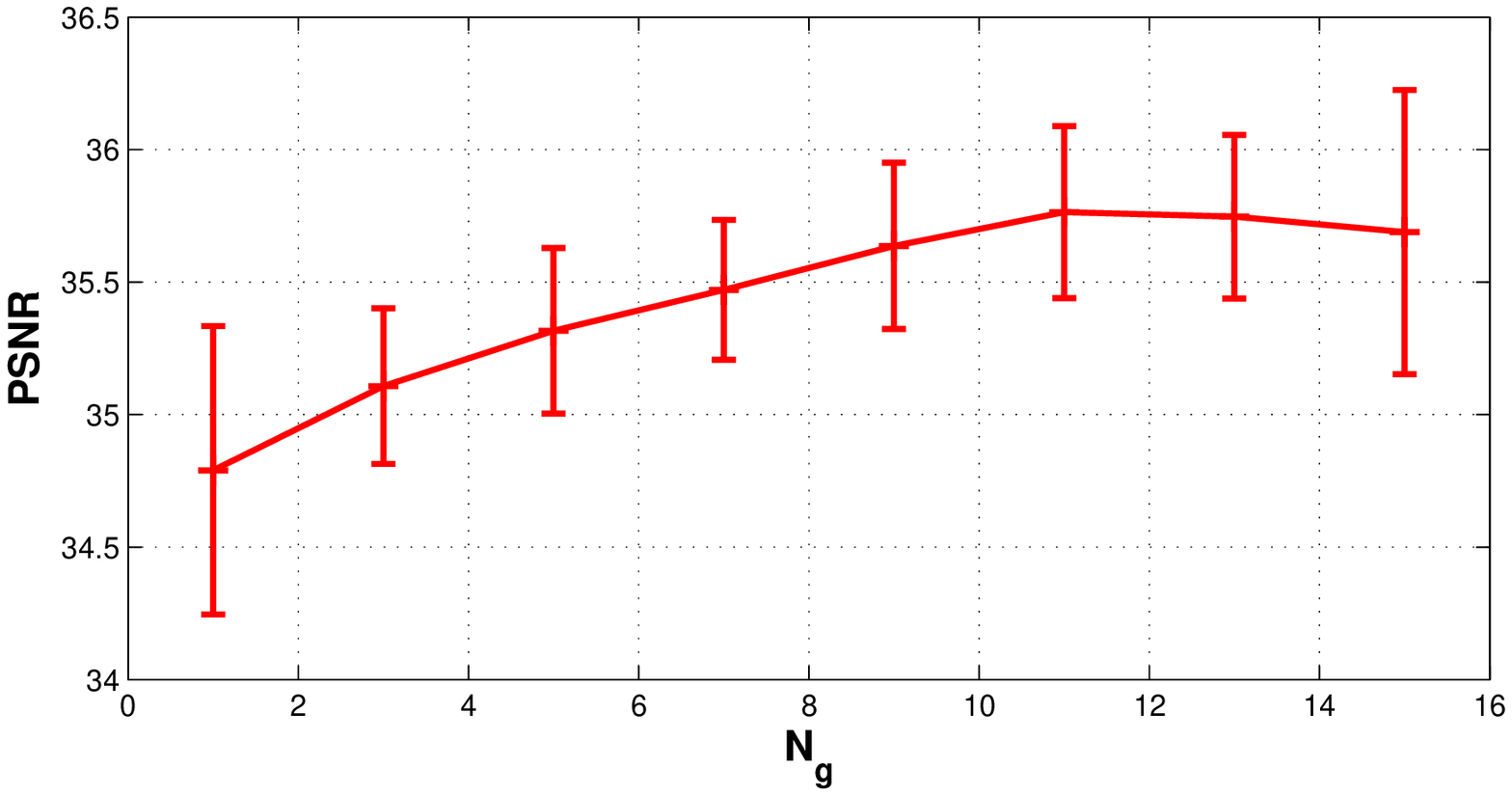}}
\subfigure[]
{\includegraphics[trim=10mm  0mm 10mm 0mm, clip=true,width=0.5\textwidth]{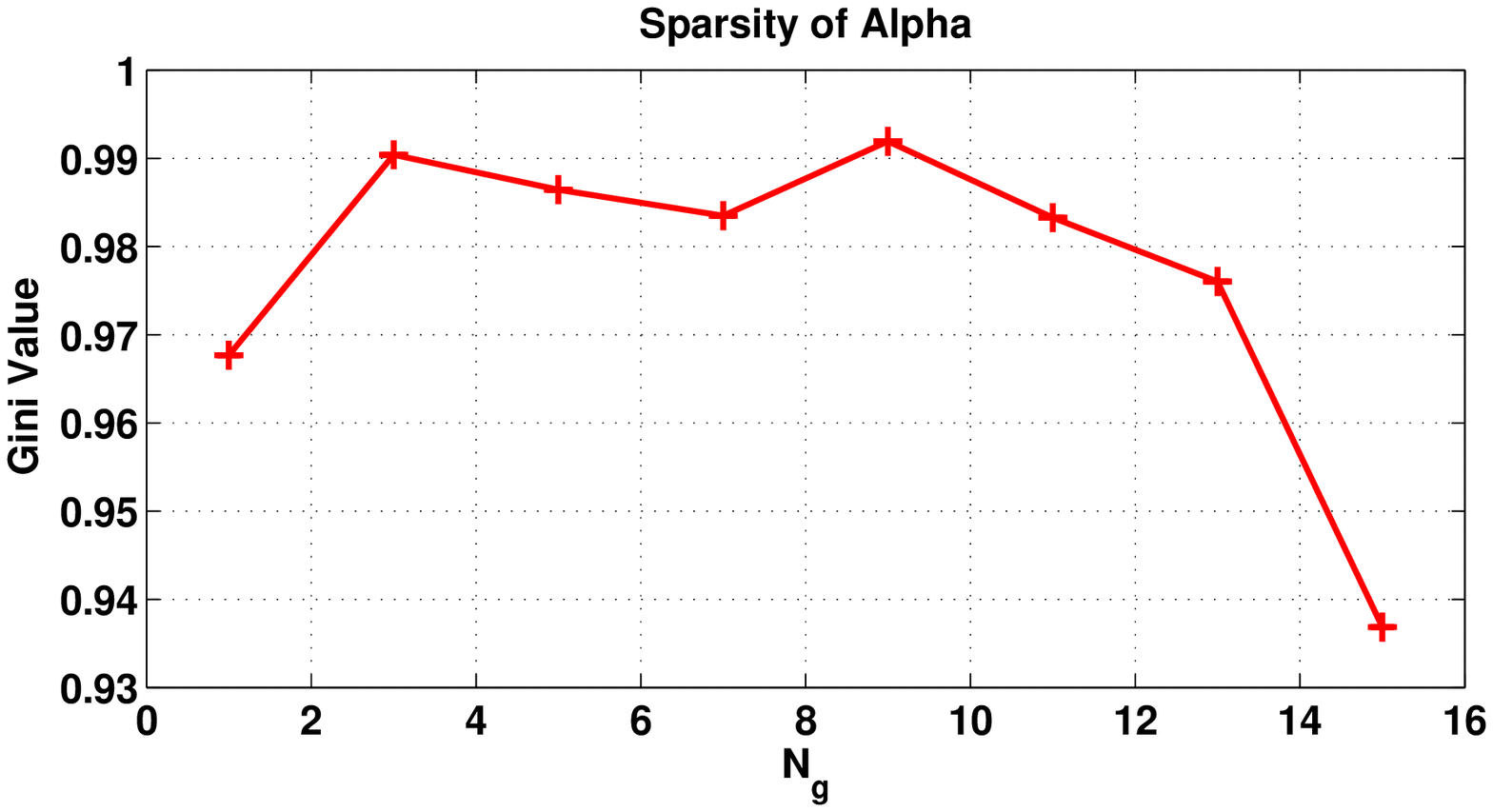}}
\caption{Influence of  group number ($N_g$) on (a) run-time of one iteration  (b) reconstruction error  (c) sparsity of $\alpha$.  }
\label{fig:Patch Group Number Sensitivity}
\end{figure}


  \paragraph{Dependence on  patch size} Figure \ref{fig:Patch Size Sensitivity} shows the sensitivity of our proposed method, to the chosen patch size, in terms of the reconstruction error.  It can be seen from the figure that increasing the patch size, beyond 6  deteriorates  the performance. This is expected since local information is not well captured when the patches are too large. On the other hand, smaller patch size results in an increased number of patches  and consequently higher computational complexity. We found $L=4$ to be a good trade-off between the computational complexity and the algorithm's performance.

\begin{figure}[h]
\centering
{\includegraphics[trim=10mm  0mm 10mm 0mm, clip=true, width=0.5\textwidth]{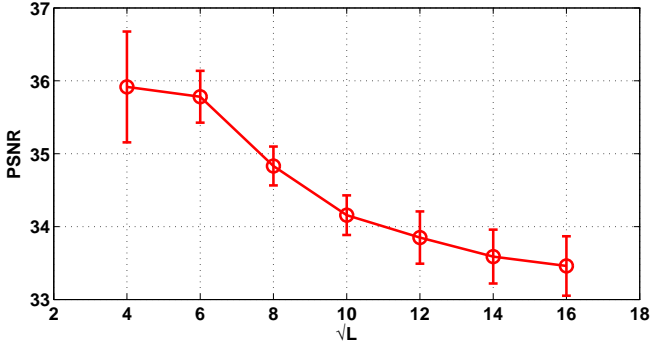}}
\caption{Reconstruction error vs. patch size ($L$).}
\label{fig:Patch Size Sensitivity}
\end{figure}

 \paragraph{Dependence on $R_1$} Figure \ref{fig:Neighbor Distance Sensitivity}  shows the reconstruction error as a function of the defined neighborhood range ($R_1$). It can be seen that our method is not sensitive to the search range and the reconstruction error only slightly decrease  with the increase of  $R_1$.  Considering that widening the search range calls for higher computational effort and also that not much reconstruction  gain is achieved, we set $R_1$ to be 13.
\begin{figure}[h]
\centering
{\includegraphics[trim=05mm  0mm 10mm 0mm, clip=true, width=0.5\textwidth]{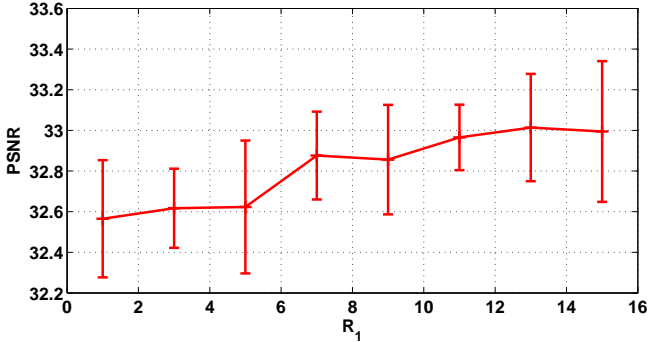}}
\caption{Reconstruction error vs. neighborhood distance ($R_1$).}
\label{fig:Neighbor Distance Sensitivity}
\end{figure}



%
%


\kkk
\begin{figure}[h]
\centering
\subfigure[]
{\includegraphics[trim=100mm 40mm 100mm 20mm, clip=true, width=0.2\textwidth]{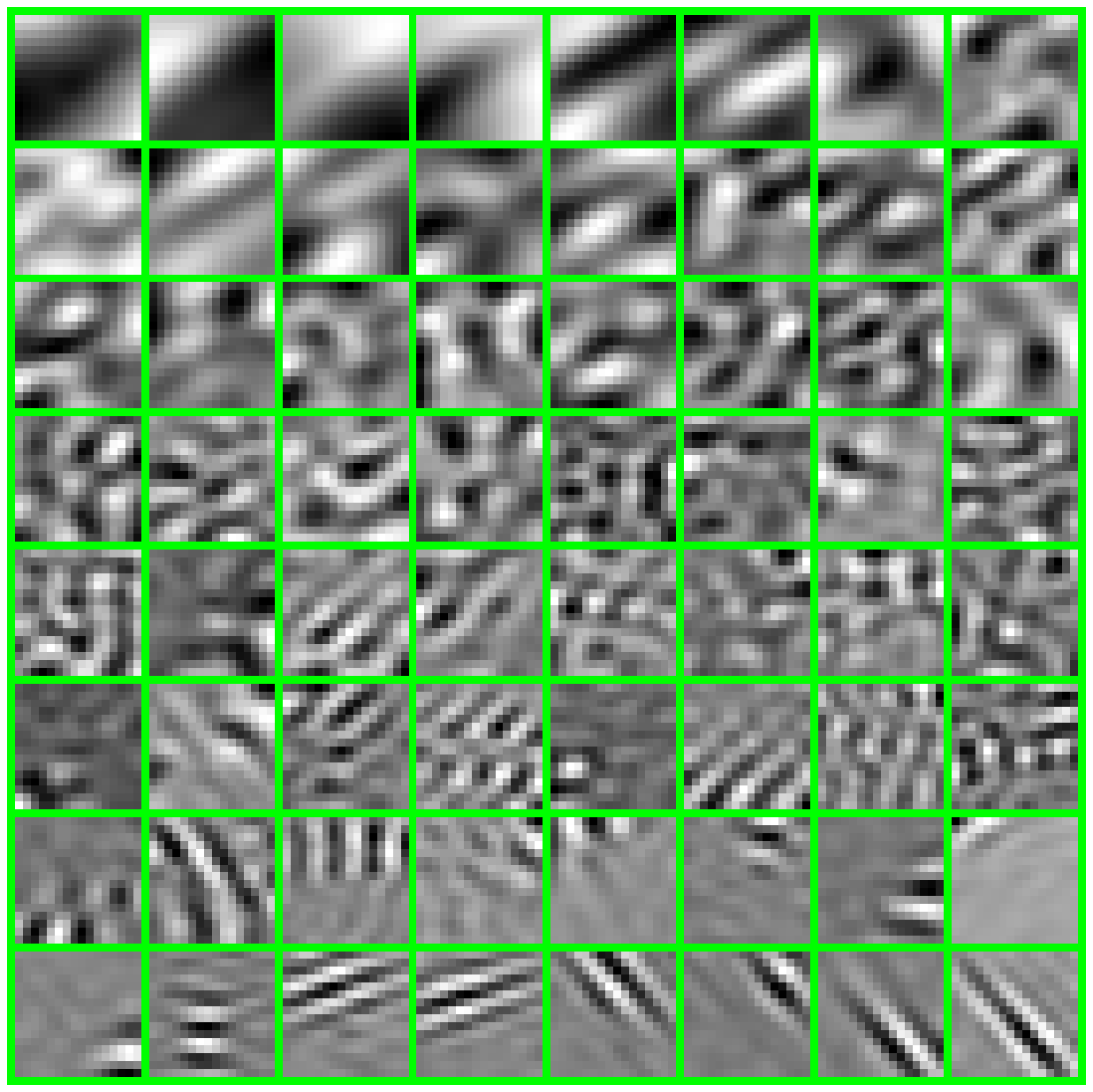}}
\subfigure[]
{\includegraphics[trim=100mm 40mm 100mm 20mm, clip=true, width=0.2\textwidth]{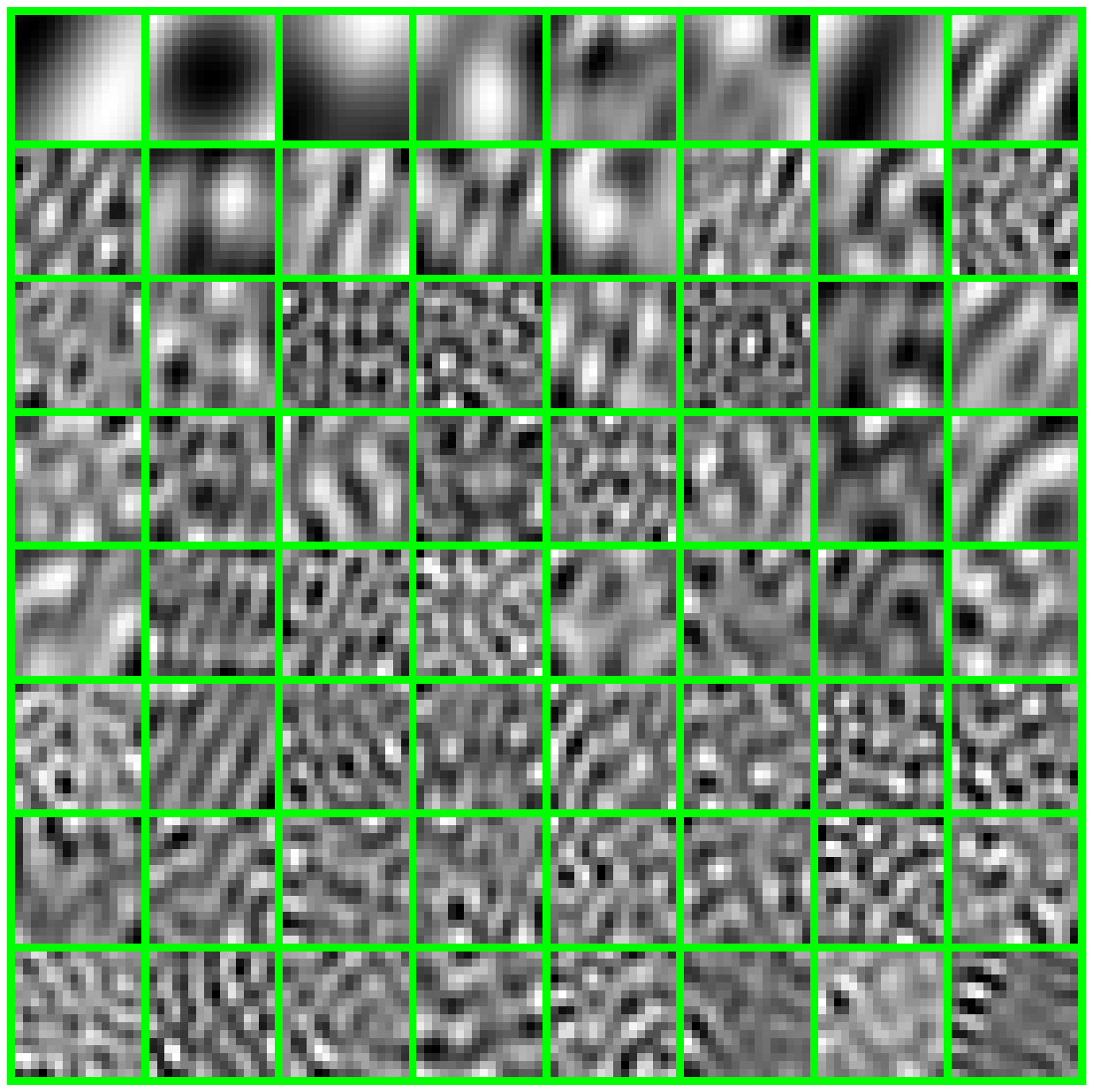}}
\subfigure[]
{\includegraphics[trim=100mm 40mm 100mm 20mm, clip=true, width=0.2\textwidth]{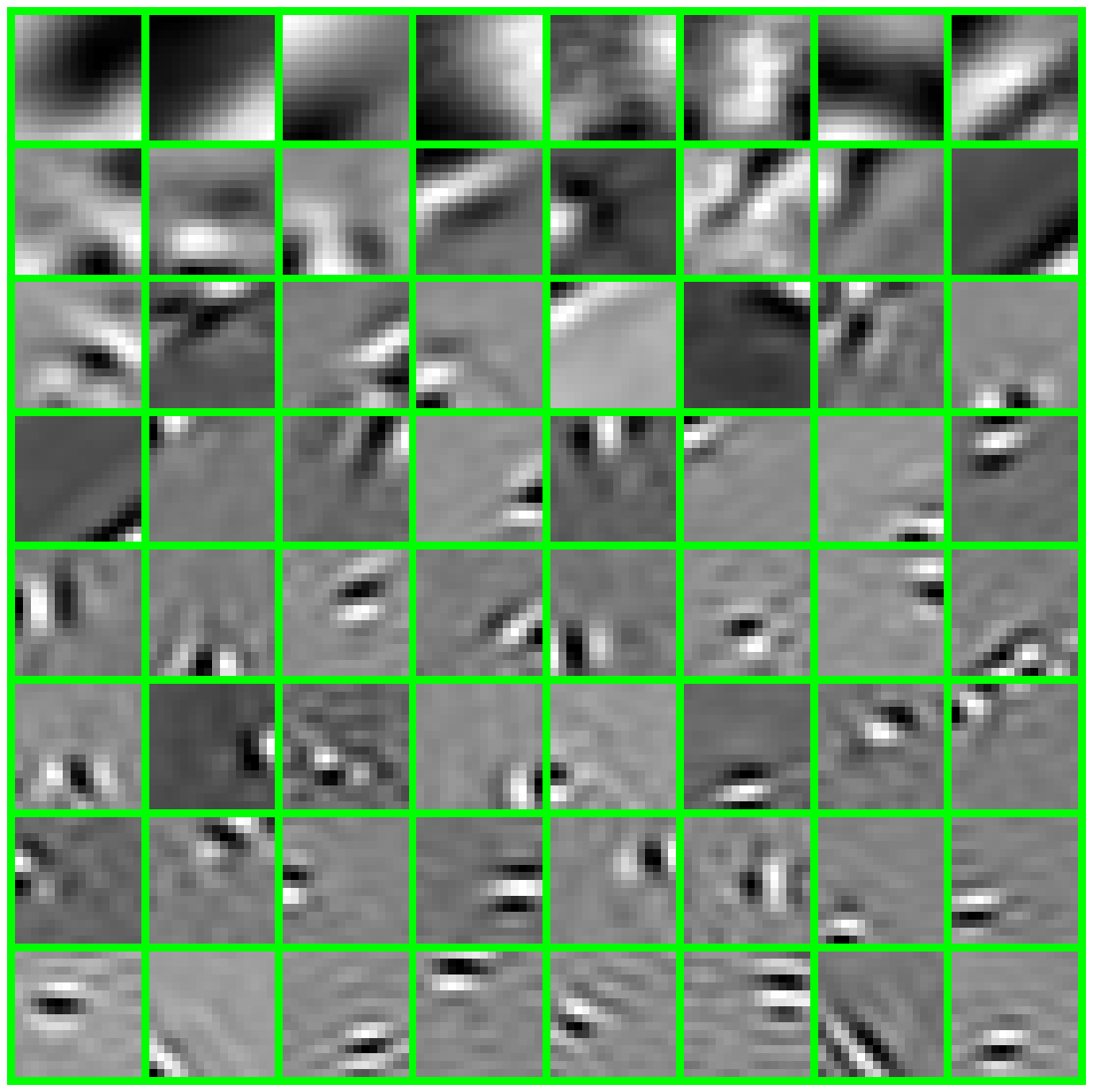}}
\caption{Trained dictionaries of 3 patch groups for the Cardiac data set at sample rate of 0.2. }
\label{fig:dic}
\end{figure}

\subsection{Summary and Future work}
Through several experiments we demonstrated the superiority of the proposed DNBG method  in terms of both quantitative and qualitative reconstruction results and also the reduced computational complexity which is due to the grouping of the patches.  Patch grouping results in significantly smaller number of patches to be handled for each class. Moreover, since the patches in each class are closely related, trained dictionaries (see Figure \ref{fig:dic}) are more tailored to that specific group. This is also evident from sparsity of $\alpha$, which indicates that a smaller number of dictionary atoms are needed to represent a patch. In addition, in contrast to  BCS that only exploits  the temporal correlations, our method uses both the spatial and temporal correlations adaptively through the learned dictionaries.

While our proposed method is developed for the single-coil MRI modality, it is possible to incorporate ''parallel imaging'' \cite{Ye2011} into the proposed method by changing the $k$-space encoding matrix to a sensitivity encoding one. It should be noted that acquiring accurate coil sensitivities for  time frame can be challenging \cite{Caballero2012}. This will be investigated in our future study.

\section{Conclusion}

\label{sec:con}

In this paper, we presented a dictionary learning based approach
which enables a detailed and fast reconstruction of the dynamic MRI
images from highly undersampled k-space. Our proposed method has three major improvements
over the state-of-the-art dictionary-based method.  Firstly, it combines both the global and patch-based sparsity.   Secondly it uses patch grouping for the DL process to reduce  computational complexity and also to train dictionaries that are more structured.
 Lastly, it uses temporal and spatial correlation of the patches and encourages close ones to employ similar dictionary settings.
 Through extensive experiments,  our proposed method has been shown to consistently achieve superior reconstruction quality, in terms of PSNR  and visual quality, with much less computational complexity than the state-of-the-art methods.
\section{Appendix}
The parameters of the DNBG method, $(\bf D_j^{(t)},z_i,s_i,\epsilon_i,\pi,\eta,\gamma_s,\gamma_e)$, are learned using Markov Chain Monte Carlo (MCMC) algorithm.
We adopted the Gibbs sampling update equations stated in \cite{Huang2014} and \cite{Zhou2011} and extended them for the DNBG model as follows:

\begin{align}
	\D_j^{(t)}&=\X_j^{(t)}\alphaj ^T(\alphaj \alphaj ^T + (L/\gamma_\varepsilon)I_K)^{-1}+E,\\\nonumber
	E_l,:&\overset{ind}\sim \N(0,(\gamma_\varepsilon\alphaj \alphaj ^T+LI_K)^{-1}), \mspace{8mu} l=1,\dots,L.
\end{align}

where $\X_j^{(t)}=\left[\P_i\xt\right]_{i\in G_j}$ and $\alphaj=\left[\alpha_i\right]_{i\in G_j}$ and $E_l$
is the $l^{th}$ column of the $E$.cite

\begin{align}
	p_{ik} &\varpropto \pi_k(1+(\gamma_\varepsilon/\gamma_{sk})d_k^Td_k)^{-\frac{1}{2}} \nonumber \\&\times\exp\left\{\frac{\gamma_\varepsilon}{2}(d_k^Tr_{i,-k})^2/(\gamma_{sk}/ \gamma_\varepsilon+d_k^Td_k)\right\},\\
	1-p_{ik} &\varpropto 1-\pi_k
\end{align}

where $r_{i,-k}$ is the error of computing the ith patch $(\P_i\x)$ with all dictionaries of DNBG except kth dictionary element, $r_{i,-k} = \P_i\x - \sum_{l \ne k} s_{il}z_{il}d_l$

\begin{align}
	&s_{ik}|z_{ik}\sim N\left( z_{ik}\frac{d_k^Tr_{i,-k}}{\gamma_\varepsilon/\gamma_{sk}+d_k^Td_k} , (\gamma_{sk}+\gamma_\varepsilon z_{ik}d_k^Td_k)^{-1}\right).
\end{align}

\begin{align}
	\gamma_\varepsilon &\sim Gamma\left(g_0+\textstyle\frac{1}{2}LN,h_0+\frac{1}{2}\sum_{i\in G_j}||\P_i\xt-\D_j^{(t)}\alpha_i||_2^2\right),\\
	\gamma_{sk }&\sim Gamma\left(e_0+ \textstyle\frac{1}{2}\sum_{i\in G_j} z_{ik}, f_0+\frac{1}{2}\sum_{i\in G_j} z_{ik}s_{ik}^2\right).
\end{align}


\begin{align*}
 \pi_{lk}^*  \sim \textrm{Beta} \left(  c_1 \eta_k + \sum_{i : \{  \| l- i \|_2 \le R_1 \}} z_{ik} \; , \; c_1(1-\eta_k) +
 \sum_{i : \{    \| l- i \|_2 \le R_1\}} (1-z_{ik})  \right)
\end{align*}

\begin{align*}
& u_{k} \sim \textrm{Unif} \left(  0 , \eta_k^{c_0 \eta_0 -1} \right)  \\
& w_{k} \sim \textrm{Unif} \left(  0 , sin^N (\pi  \eta_k ) \right)  \\
& v_{k} \sim \textrm{Unif} \left(  0 , (1-\eta_k)^{c_0 (1- \eta_0) -1} \right)
\end{align*}

\begin{align*}
& \eta_{k} \sim \textrm{Exp} \left(  - c_1 \sum_{l=1}^N \log \left(  \frac{\pi^*_{lk}}{1 - \pi^*_{lk} } \right)  \right)  \textrm{I} (\eta_k)
\end{align*}

\ifCLASSOPTIONcaptionsoff
  \newpage
\fi

\bibliographystyle{IEEEbib}
\bibliography{references}

\end{document}